\documentclass[letterpaper, 10pt, conference]{ieeeconf}
\IEEEoverridecommandlockouts
\usepackage{graphicx}
\usepackage{amsmath,amssymb}
\usepackage{booktabs}
\usepackage{hyperref}
\usepackage{cite}
\usepackage{xcolor}
\usepackage{multirow}

\title{\LARGE \bf Safe Aerial 3D Path Planning for Autonomous\\ UAVs using Magnetic Potential Fields}
\author{Haechan Mark Bong$^{1, 2}$,
        Giovanni Beltrame$^{1, 2}$
        \thanks{$^{1}$Department of Computer Engineering and Software Engineering, Polytechnique Montréal, QC., Canada. {\tt\small haechan.bong@etud.polymtl.ca}}
        \thanks{$^{2}$MILA, QC., Canada.}
}

\begin{document}
\bstctlcite{bstctl:etal}
\maketitle
\thispagestyle{empty}
\pagestyle{empty}

\begin{abstract}
Safe autonomous Uncrewed Aerial Vehicle (UAV) navigation in urban environments requires real-time path planning that avoids obstacles. MaxConvNet is a potential-field planner that leverages properties of Maxwell's equations to generate a path to the goal without local minima. We extend the 2D MaxConvNet magnetic field planner to 3D, using a convolutional autoencoder to predict obstacle-aware potential fields from LiDAR-derived $101^3$ voxel grids. Evaluation across 100 randomized closed-loop trials in two distinct Cosys-AirSim urban environments, a dense night-time cityscape and a suburban district shows a 100\% path planning success rate on both maps without retraining. In offline path planning, 3DMaxConvNet produces path lengths comparable to A* on unseen maps while reducing runtime from $0.155$--$0.17$s to $0.087$--$0.089$s, or about $1.7$--$1.95$ times faster than A*. Against RRT*(3k), 3DMaxConvNet achieves similar path quality while reducing planning runtime from $17.2$--$17.5$s to about $0.09$s, which is roughly $193$--$201$ times faster than RRT*(3k).
\end{abstract}

\section{Introduction}

Safe path planning is a core requirement for deploying autonomous UAVs in urban environments. Classical artificial potential field (APF) methods~\cite{khatib1986potential} offer elegant gradient-based planning but suffer from local minima that can trap the UAV indefinitely, which is a critical safety concern. Sampling-based planners like RRT*~\cite{karaman2011samplingbasedalgorithmsoptimalmotion} provide asymptotic optimality, but they are too slow for real-time replanning. Graph-based methods such as A*~\cite{hart1968} guarantee optimality on discrete grids but scale poorly to high-resolution 3D volumes.

Recent learning-based approaches have shown promise for fast path planning. NTFields~\cite{ntfields2024} uses neural implicit representations to learn time-optimal trajectories, while physics-informed neural networks~\cite{pinn_review} approximate PDE solutions for planning. 2DMaxConvNet~\cite{moussa2021} demonstrated that a 2D convolutional autoencoder can approximate Maxwell's steady-state conduction equation in real-time on 2D grid environments. 2DMaxConvNet is a potential-field planner that uses properties of Maxwell's equations to produce a path to the goal without local minima. However, their work was limited to 2D grids with synthetic obstacles and did not address the safety challenges of real 3D sensor data. Additionally, their environment setup considered only simple geometric obstacles (e.g., rectangles), which does not capture the complexity of real urban environments.

\begin{figure}[!t]
    \centering
    \includegraphics[width=\columnwidth]{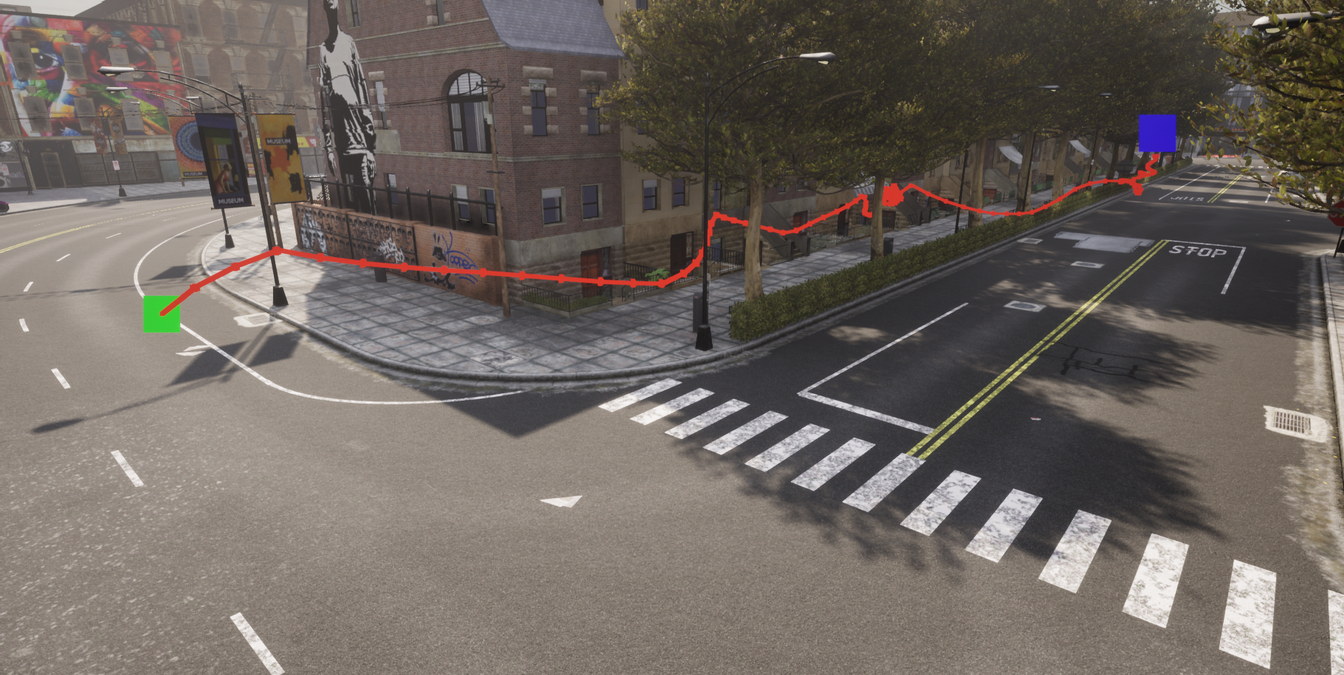}
    \caption{Example path: the red curve is the path following a magnetic field, and the blue and green squares are the start and goal points, respectively.}
    \label{fig:example_path}
\end{figure}

\begin{figure}[!t]
    \centering
    \includegraphics[width=\columnwidth]{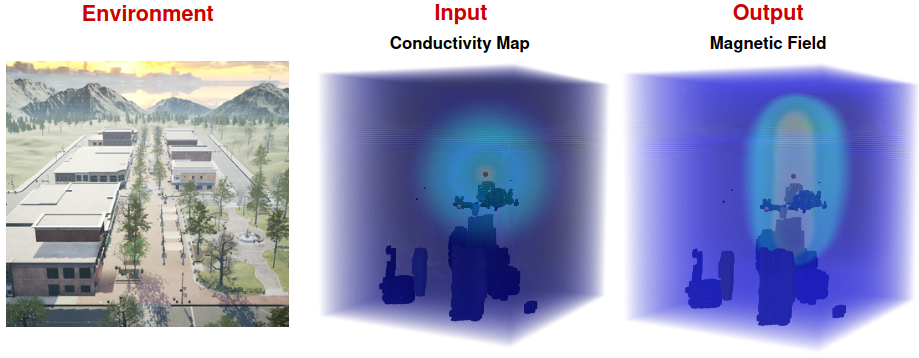}
    \caption{3DMaxConvNet training map: Downtown West~\cite{downtownwest}, used for collecting training scenes.}
    \label{fig:train_env}
\end{figure}

We extend this approach to 3D urban UAV navigation with LiDAR data, targeting safe aerial operations in 3D realistic (cluttered) environments. Figure~\ref{fig:example_path} shows an example of a path that follows magnetic fields from start to goal. Figure~\ref{fig:train_env} shows the computed conductivity and magnetic field of a 3D environment~\cite{downtownwest} that represents an urban town with natural effects (e.g. sun light, wind, etc.) to simulate a real environment. Our main contributions include: (1)~extension of MaxConvNet from 2D to 3D with real LiDAR sensor data; (2) evaluation across 100 closed-loop trials on two unseen complex 3D environments.

\begin{figure*}[!t]
    \centering
    \includegraphics[width=\textwidth]{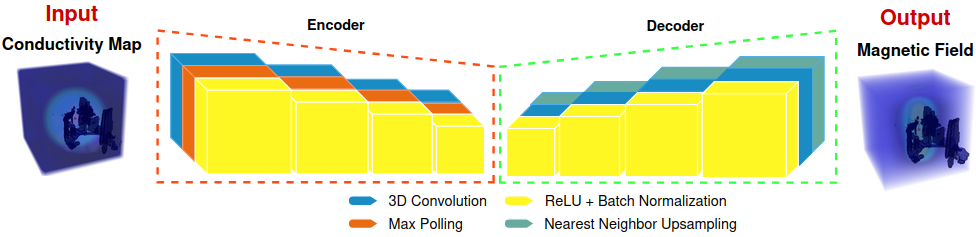}
    \caption{3DMaxConvNet architecture. A conductivity image of the environment is first processed by an encoder consisting of five stages of Conv3D, ReLU, and max-pooling. A decoder with five stages of upsampling, Conv3D, and ReLU then reconstructs the magnetic field. Dropout and $L_2$ regularization are used to reduce overfitting. Ground-truth magnetic field distributions are computed from conductivity maps collected in the Downtown West~\cite{downtownwest} Unreal Engine environment and Cosys-AirSim~\cite{shah2018airsim} environment to define the training loss.}
    \label{fig:architecture}
\end{figure*}

\section{Related Work}

\textbf{Potential field methods.} Khatib~\cite{khatib1986potential} introduced APFs for real-time obstacle avoidance, combining attractive goal fields with repulsive obstacle fields. While computationally efficient, APFs suffer from local minima in narrow passages and between closely spaced obstacles~\cite{koren1991}. Navigation functions~\cite{rimon1992} can provide convergence guarantees under specific assumptions, but they require global environment knowledge and that is difficult to construct in complex 3D spaces.

\textbf{Sampling-based planners.} RRT*~\cite{karaman2011samplingbasedalgorithmsoptimalmotion} provides asymptotic optimality through incremental rewiring but convergence is slow in high-dimensional spaces. Informed RRT*~\cite{gammell2014} improves convergence by focusing on sampling, yet remains orders of magnitude slower than real-time requirements for iterative replanning during flight.

\textbf{Learning-based planning.} Neural motion planners~\cite{qureshi2019motionplanningnetworks} learn to generate paths from point cloud observations. NTFields~\cite{ntfields2024} encode time-optimal navigation as a neural implicit field trained using the Eikonal equation. Value iteration networks~\cite{tamar2017valueiterationnetworks} embed differentiable planning modules within neural networks. MaxConvNet~\cite{moussa2021} uniquely frames planning as electromagnetic field prediction, leveraging the physics which guarantees no local maxima in the magnetic field given a path exists.

\textbf{Safe aerial navigation.} Ensuring safety during autonomous flight requires both planning-level guarantees (collision-free paths) and execution-level robustness (reactive avoidance). Recent work combines learned planners with safety filters~\cite{tordesillas2021} or control barrier functions for runtime safety. Our approach addresses safety through multi-layered defense: obstacle-aware magnetic field prediction that theoretically guarantees no local maxima. This theory is a critical safety component where if a feasable path exists between start and goal positions, by following the magnetic field, the path will be found towards the goal.

\section{Method}

\subsection{Conductivity Field Formulation}

The environment is modeled as an electromagnetic conduction problem on a $101^3$ voxel grid ($\pm$100m, 2m resolution). Obstacle voxels (from LiDAR, dilated by 4 voxels for safety margin) have zero conductivity ($\sigma_o\!=\!0$), the goal has high conductivity ($\sigma_g\!=\!10^6$), and free space combines initial conductivity ($\sigma_i\!=\!10^2$) to determine the distance-weighted conductivity as defefined below:
 $$\sigma_e(x, y, z) = \frac{\sigma_g}{\sqrt{(x - x_g)^2 + (y - y_g)^2 + (z - z_g)^2}} $$

\subsection{3DMaxConvNet Architecture}

We extend MaxConvNet~\cite{moussa2021} from $N^2$ grids to $101^3$ volumes. The original 2D kernels ($11\!\times\!11$ to $3\!\times\!3$) would yield $11^3\!=\!1331$ weights per filter in 3D, resulting in 6.25M parameters and ${\sim}$28s inference. We replace all kernels with $3\!\times\!3\!\times\!3$ (27 weights) and use nearest-neighbor upsampling in the decoder, reducing parameters to 4.1M and inference to 95ms (GPU). Five pooling stages yield an effective receptive field of 94 voxels, covering the full domain.

\textbf{Training:} 749 scenes collected from the Downtown West~\cite{downtownwest} Cosys-AirSim~\cite{shah2018airsim} environment (32,768-point LiDAR scans). Ground truth was solved using SciPy's conjugate gradient solver~\cite{2020SciPy-NMeth} (${\sim}$10s/scene). Trained 200 epochs with AdamW ($\text{lr}\!=\!10^{-3}$), cosine annealing, MSE loss. Validation $L_2$ error: 0.103.

\subsection{Path Optimization and Safety Layers}

\textbf{Gradient smoothing.} The predicted field's gradients are smoothed with a 3D Gaussian filter ($\sigma\!=\!5$ voxels) to suppress noise from sparse LiDAR, eliminating path oscillation in open space.

\textbf{Momentum gradient ascent.} Unlike the original ADAM optimizer~\cite{kingma2015} which normalizes each gradient component independently (amplifying noise in low-gradient directions), we use momentum-only gradient ascent ($\beta_1\!=\!0.9$, lr$\!=\!0.1$) that preserves the gradient's directional structure.

\section{Experiments}

\begin{figure*}[!t]
    \centering
    \includegraphics[width=\textwidth]{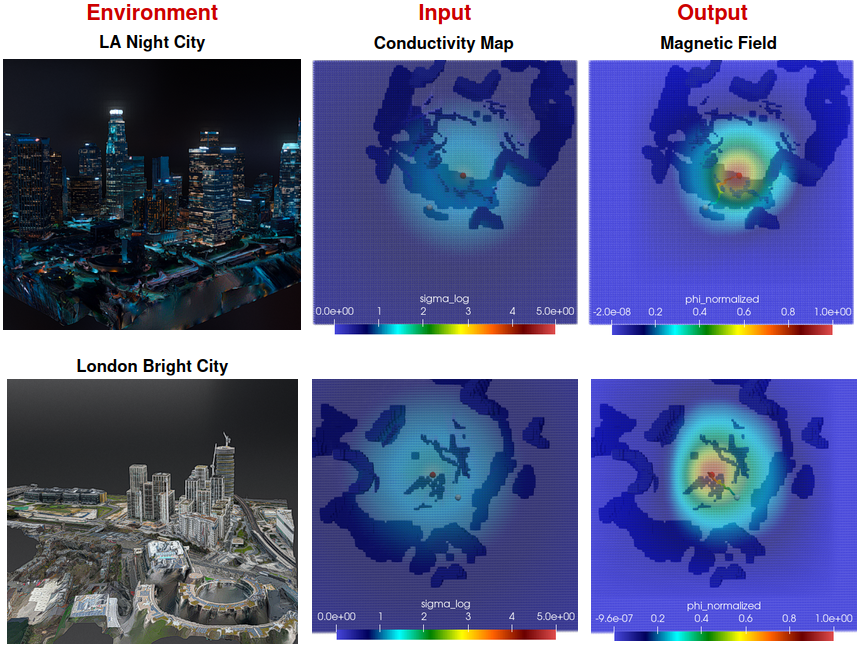}
    \caption{Field visualizations using Cosys-AirSim~\cite{shah2018airsim}, with LA Night City~\cite{lanightcity} (top) and London White City~\cite{londonwhitecity} (bottom), extracted from experiment 45. \textbf{Conductivity:} The raw conductivity field ($\sigma$) and its log-scaled representation ($\sigma_{\log}$) showing the distance-weighted gradient from free space toward the goal. \textbf{Magnetic field:} The 3DMaxConvNet predicted magnetic field ($\phi$) and its normalized form ($\phi_{\mathrm{normalized}}$) revealing the gradient structure that guides the UAV toward the goal. The white and red sphere marks the start, the goal, respectively and the planned path follows the field gradient.}
    \label{fig:environment}
\end{figure*}

We evaluate in Cosys-AirSim~\cite{shah2018airsim} on two urban environments (Fig.~\ref{fig:environment}):

\textbf{LA Night City}~\cite{lanightcity} is a dense urban environment, mirroring  the real LA city by featuring high-rise buildings, narrow streets, and night-time lighting conditions. The complex geometry with tall structures and confined passages tests the planner's ability to navigate through cluttered 3D obstacle fields.

\textbf{London White City}~\cite{londonwhitecity} is a suburban district with low-rise buildings and open spaces. The sparser obstacle layout tests generalization to environments with different obstacle density and distribution than the training data.

Each trial places start and goal 20--50m apart at 3--15m altitude, validated collision-free with a 10m clearance check. The low altitude is intentional since high altitudes does not not have dense obstacles. The UAV has a 32,768-point LiDAR sensor with 100m range. Maximum 100 planning iterations per trial. Figure~\ref{fig:environment} shows representative field visualizations for both LA Night City and London White City.

\subsection{Results}

We evaluate in two modes, 50 experiments per map (Table~\ref{tab:results}). \textbf{Offline:} given a single pre-built obstacle grid (1 path planning iteration), A*~\cite{hart1968}, RRT*(3k)~\cite{karaman2011samplingbasedalgorithmsoptimalmotion}, and 3DMaxConvNet each compute a single path from start to goal. This evaluation uses no UAV, no flight, and no LiDAR. All methods receive the same grid, isolating planning quality and speed. Path lengths are in voxel units (times 2m for physical distance). \textbf{Closed-loop:} the UAV flies in Cosys-AirSim, iteratively capturing LiDAR points, rebuilding the grid, replanning, and flying segments. The start to goal distance is randomly distributed between 50m to 80m. The closed-loop path is typically longer because the UAV has limited visibility (100m LiDAR range), replans as it discovers new obstacles. 

\begin{table}[t]
\centering
\caption{Path planning evaluation. Top: offline path quality on held-out scenes from each environment (path length in voxel units times 2\,m). Bottom: closed-loop flight performance (50 trials each, no retraining).}
\label{tab:results}
\setlength{\tabcolsep}{3.5pt}
\footnotesize
\begin{tabular}{@{}llcc@{}}
\toprule
\multicolumn{4}{c}{Offline Path Quality (1 Iteration of Planning)} \\
\midrule
\textbf{Environment} & \textbf{Method} & \textbf{Path (m)} & \textbf{Runtime (s)} \\
\midrule
\multirow{3}{*}{\shortstack[l]{LA Night\\City}}
  & A* & $54.6 \pm 17.8$ & $0.155 \pm 0.16$ \\
  & RRT*(3k) & $57.5 \pm 19.6$ & 17.2 \\
  & 3DMaxConvNet & $\mathbf{58.5 \pm 17.8}$ & \textbf{0.089} \\
\midrule
\multirow{3}{*}{\shortstack[l]{London\\White City}}
  & A* & $57.6 \pm 19.8$ & 0.17 \\
  & RRT*(3k) & $59.8 \pm 21.8$ & 17.5 \\
  & 3DMaxConvNet & $\mathbf{59.3 \pm 19.4}$ & \textbf{0.087} \\
\midrule
\multicolumn{4}{c}{Closed-Loop Flight (Start to Goal)} \\
\midrule
\textbf{Environment} & \textbf{Method} & \textbf{Path (m)} & \textbf{Planning} \\
\midrule
\multirow{2}{*}{\shortstack[l]{LA Night\\City}}
  & A* & $81.7 \pm 68.4$ & $25.2 \pm 25.9$ \\
  & 3DMaxConvNet & $109.0 \pm 60.2$ & $31.0 \pm 20.5$ \\
\midrule
\multirow{2}{*}{\shortstack[l]{London\\White City}}
  & A* & $73.4 \pm 8.1$ & $20.6 \pm 3.3$ \\
  & 3DMaxConvNet & $74.1 \pm 10.6$ & $19.9 \pm 2.8$ \\
\bottomrule
\end{tabular}
\end{table}

\textbf{Offline path quality.} Table~\ref{tab:results} shows that 3DMaxConvNet produces path lengths comparable to those of classical planners while remaining substantially faster. On LA Night City, 3DMaxConvNet yields a slightly longer mean path than A* ($58.5$m vs. $54.6$m) but reduces runtime from $0.155$s to $0.089$s. On London White City, 3DMaxConvNet is close to A* in path length ($59.3$m vs. $57.6$m) while reducing runtime from $0.17$s to $0.087$s. RRT*(3k) is the slowest method on both maps, requiring $17.2$s--$17.5$s per plan, which is roughly two orders of magnitude slower than 3DMaxConvNet.

\textbf{Field accuracy.} The relative $L_2$ error between the predicted and ground-truth fields in London White City is $L_2\!=\!0.051$ (5.1\%). This is lower than the error in LA Night City, which reaches $L_2\!=\!0.153$ (15.3\%), reflecting the challenge of dense structures not seen during training. Even with 15\% field error, the \emph{gradient direction} remains accurate for path planning (100\% success rate) because the Gaussian smoothing averages over local field variations.

\textbf{Generalization.} 3DMaxConvNet achieves 100\% closed-loop success on both unseen environments without retraining (Table~\ref{tab:results}, bottom). In London White City, 3DMaxConvNet produces a flown path (start to goal) comparable to A* ($74.1 \pm 10.6$m vs. $73.4 \pm 8.1$m) while requiring slightly fewer planning steps on average ($19.9 \pm 2.8$ vs. $20.6 \pm 3.3$). In LA Night City, the denser structure increases the difficulty for both methods; 3DMaxConvNet produces a longer flown path than A* ($109.0 \pm 60.2$m vs. $81.7 \pm 68.4$m) and uses more planning steps ($31.0 \pm 20.5$ vs. $25.2 \pm 25.9$), but every trial still reaches the goal without collision.

\textbf{Timing.} In the offline evaluation, 3DMaxConvNet plans in $0.087$s--$0.089$s for both unseen maps, compared with $0.155$s--$0.17$s for A* and $17.2$--$17.5$s for RRT*(3k). This makes 3DMaxConvNet about $1.7$--$1.95$ times faster than A* and roughly $193$--$201$ times faster than RRT*(3k), while maintaining similar path quality. These results indicate that the 3DMaxConvNet path planner is fast enough for iterative replanning in closed-loop flight.

\section{Discussion}

\begin{figure}[!t]
    \centering
    \includegraphics[width=\columnwidth]{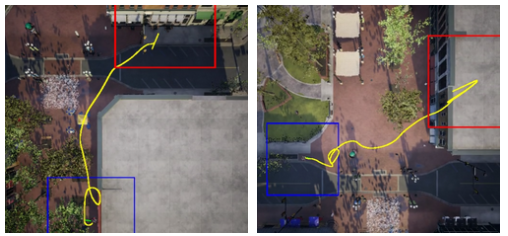}
    \caption{Bird's-eye view (BEV) of non-optimal (circular) UAV trajectories. Yellow denotes the flown path; blue and red squares indicate start and goal areas, respectively. Larger square size corresponds to higher altitude.}
    \label{fig:bad_paths}
\end{figure}

\textbf{Limitations.} As shown in Fig.~\ref{fig:bad_paths}, the UAV always arrives at the goal, but it does not always follow the optimal path and can exhibit curly or circular motion. This behavior occurs because the reactive collision-avoidance layer from AirSim can drift the UAV away from the planned magnetic-field path when an obstacle is near, and the policy is sensitive to moving leaves from trees under wind. The system also handles only static environments. Also, 3DMaxConvNet predicts from a single LiDAR snapshot with no velocity estimation. The reactive avoidance layer (3m range, repulsive only) cannot anticipate head-on collisions (dynamic obstacles).

\textbf{Future work.} We aim to reduce inference latency through model compression to close the gap with the 2DMaxConvNet's real-time rates. In addition, the key direction is extending to dynamic obstacle environments through obstacle tracking and predictive collision avoidance. We plan to validate on real outdoor platforms where sensor noise, wind disturbances, and unpredictable agents present challenges beyond simulation.

\bibliographystyle{IEEEtran-custom}
\bibliography{refs}

\end{document}